\documentclass[letterpaper, 10 pt, conference]{ieeeconf}  

\IEEEoverridecommandlockouts                              

\usepackage{graphicx}
\usepackage{epstopdf}
\usepackage{epsfig} 
\usepackage{amsmath} 
\usepackage{amssymb}  
\usepackage{mathtools}
\usepackage{pgfplots}
\usepackage{amsfonts}
\usepackage{url}
\usepackage[]{units}
\usepackage{tikz}
\usepackage{multirow}
\usepackage{multicol}
\usepackage{lipsum} 
\usetikzlibrary{calc}
\usetikzlibrary{positioning}
\graphicspath{{./img/}}
\newcommand\Tstrut{\rule{0pt}{2.6ex}}  

\title{\LARGE \bf Vision-Based Proprioceptive Sensing for Soft Inflatable Actuators}

\author{Peter Werner, Matthias Hofer, Carmelo Sferrazza and Raffaello D'Andrea
\thanks{The authors are members of the Institute for Dynamic Systems and Control, ETH Z\"urich, Switzerland. Email correspondence to Peter Werner {\tt\small wernerpe@ethz.ch}.}}%
\begin{document}

\maketitle
\thispagestyle{empty}
\pagestyle{empty}
\begin{abstract}\label{sec:Abstract}
This paper presents a vision-based sensing approach for a soft linear actuator, which is equipped with an integrated camera. The proposed vision-based sensing pipeline predicts the three-dimensional position of a point of interest on the actuator. To train and evaluate the algorithm, predictions are compared to ground truth data from an external motion capture system. An off-the-shelf distance sensor is integrated in a similar actuator and its performance is used as a baseline for comparison. The resulting sensing pipeline runs at 40 Hz in real-time on a standard laptop and is additionally used for closed loop elongation control of the actuator. It is shown that the approach can achieve comparable accuracy to the distance sensor.
\end{abstract}


\section{Introduction}\label{sec:Introduction}
Due to their intrinsic properties, inflatable soft robotic systems show promise in overcoming barriers encountered with classic rigid robotic systems \cite{PPolygerinos_SRR}. Soft materials provide robots with intrinsic compliance and the ability to interact with their surroundings in a safer and more resilient way. However, these soft systems typically result in a high number of degrees of freedom \cite{DRus_Review}. Furthermore, modeling dynamic behavior is challenging due to the non-linear material properties. Therefore, sensory feedback is crucial for the control of soft robots \cite{HWang_TowardsPerceptiveSR}.

Proprioception in robotics refers to sensing the robot's own internal state, and is an active field of research in soft robotic systems. A number of different approaches are investigated for retrieving the shape of a soft robot relying only on internal sensors \cite{HWang_TowardsPerceptiveSR}. In the context of optical sensors, stretchable strain sensors based on optical waveguides are employed, where the light transmission properties change when the waveguide is deformed \cite{HZhao_Waveguide}. The changing light intensity due to deflection is used in \cite{MDobrzynski_Contacless}. The idea is to attach a flexible circuit board, housing a light sensor and various photodiodes, to a soft object. When the object deforms it causes the flexible circuit board to bend and as a consequence, the light intensity to change. A similar idea is proposed in \cite{HYang_ModelAndAnalysis}, where an LED and a phototransistor are mounted on the opposing ends of an inflatable linear actuator. When the actuator is inflated, the light intensity measured by the photodiode decreases as a function of the elongation of the actuator.

\begin{figure}[t!]
\centering
\includegraphics[trim=0mm 0mm 0mm 0mm, clip, width=\columnwidth]{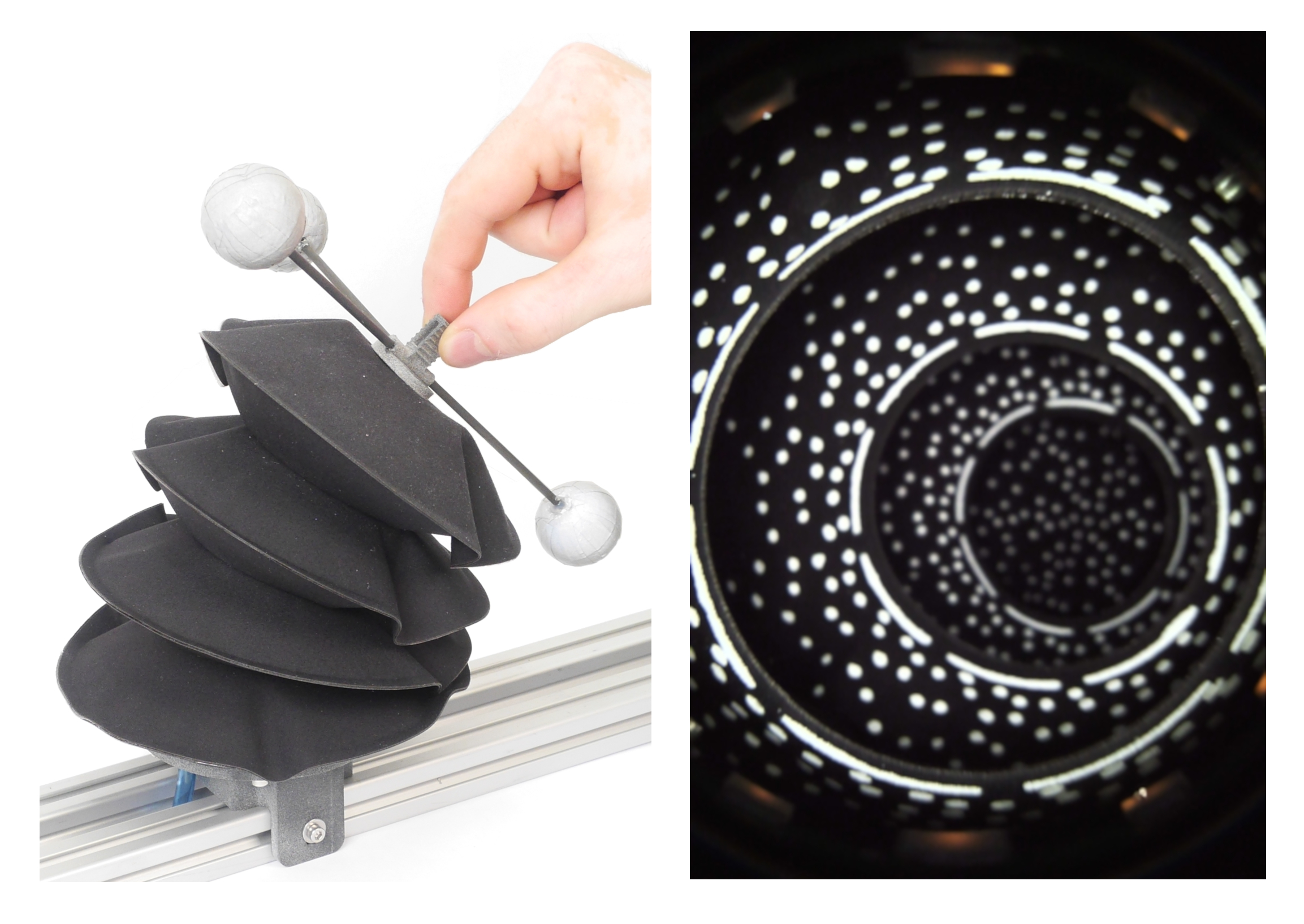}
\caption{Left: Inflatable linear soft actuator with integrated camera in perturbed state. Markers for an external motion capture system are mounted on top for the acquisition of ground truth data. Right: Image from the integrated camera showing the employed pattern. The proposed sensing method is used to predict the position of a point of interest located on the grip, using only images from the integrated camera.}
\label{Fig:SoftActuator}
\end{figure}

Vision-based sensing relying on a camera to measure the deformation of a soft material is a promising approach investigated actively in the field of tactile sensing. Rich visual information about the strain of the sensor's soft surface is provided by a camera tracking markers embedded in a soft material (\cite{CChorley_TacTip}, \cite{CSferrazza_DesignMot}) or observing the reflection of light on a deformed surface \cite{WYuan_GelSight}.

Vision-based sensing is promising because of two reasons. Firstly, the sensor (i.e. the camera) is not required to mechanically interact with the soft material of the robot. Therefore, the sensor does not need to match the compliance of the soft material employed, in order to avoid stress concentrations or a degradation of the overall compliance of the system. Secondly, vision-based approaches provide high spatial resolution and minimal wiring \cite{ZKappassov_TactileSensing}. However, the use of a camera generally leads to a bulkier structure and requires the points of interest to be in the field of view of the camera. 

An approach which combines vision-based tactile sensing with pneumatic actuation is demonstrated in \cite{BMcInroe_TowardsAS}. An internally mounted camera tracks markers attached to a soft membrane, which can be deformed by increasing the internal pressure. The authors of \cite{RWang_RealTimeSoftRobot} present the three-dimensional shape reconstruction of soft objects based on a self-observing camera. Ground truth data from two external depth cameras is used to train an artificial neural network that runs on a GPU and predicts the object shape from the self-observing camera images.

In this paper, we present an approach for sensing the state of an inflatable, fabric-based actuator relying on the integration of an RGB camera into the actuator. We present a systematic method to apply a distinctive pattern to the interior of the actuator during manufacturing. A number of computationally cheap features are extracted from the raw camera image and used as inputs to a support vector regressor (SVR), where ground truth data from a motion capture system are used to train the model. The performance of the camera-based approach is compared to an infrared time-of-flight sensor serving as the baseline for measuring the linear expansion of the soft actuator. It is shown that the camera-based approach can track the position of a point of interest on the actuator in real-time at $\unit[40]{Hz}$ with relatively high accuracy. Finally, the camera-based state prediction is used in closed loop to control the elongation of the inflatable actuator.

The remainder of this paper is organized as follows: The manufacturing of the actuator including the pattern and the integration of the camera is discussed in Section \ref{sec:hardware}. The feature extraction from the raw camera image and the applied machine learning pipeline is outlined in Section \ref{sec:method}. Experimental results of the real-time position estimation and the closed loop elongation control are presented in Section \ref{sec:experiments}. Finally, a conclusion is drawn in Section \ref{sec:conclusion}.

\section{Hardware}\label{sec:hardware}
The design and fabrication of the actuator and the test setup used in this paper are presented in this section. 

\subsection{Actuator}\label{ssec:actuator}

The actuator consists of four circular bellows with a diameter of 140 mm when collapsed. Inflating these bellows causes the actuator to expand longitudinally. A camera is integrated in the interior of the actuator and a pattern is applied to the fabric layers in its line of sight.  A small frame with three markers required for the motion capture system is glued (using Loctite 4850) to the top bellow. The assembled system can be seen in Fig. \ref{Fig:SoftActuator}. 

The actuator is manufactured using the approach described in \cite{yang2018new}. To summarize: It is built from sheets of fabric material that have a sandwich structure. This material is composed of two layers of poplin fabric (polyester cotton blend 65/35) stacked above and below a layer of thermoplastic polyurethane (TPU) film (HM65-PA, \unit[0.1]{mm} by perfectex) that are fused in a heat press. The bellows are composed of rings and a lid from the fabric material and additional TPU ring-shaped seams. These parts are all cut using a laser cutter. The bellows are constructed by stacking the fabric parts with the TPU seams in-between and fusing them sequentially in a heat press. The resulting processed fabric is inextensible. A more detailed description of the manufacturing process is given in \cite{yang2018new} (Layered Manufacturing-Type I).

\begin{figure}
	\centering
	\includegraphics[width=\columnwidth]{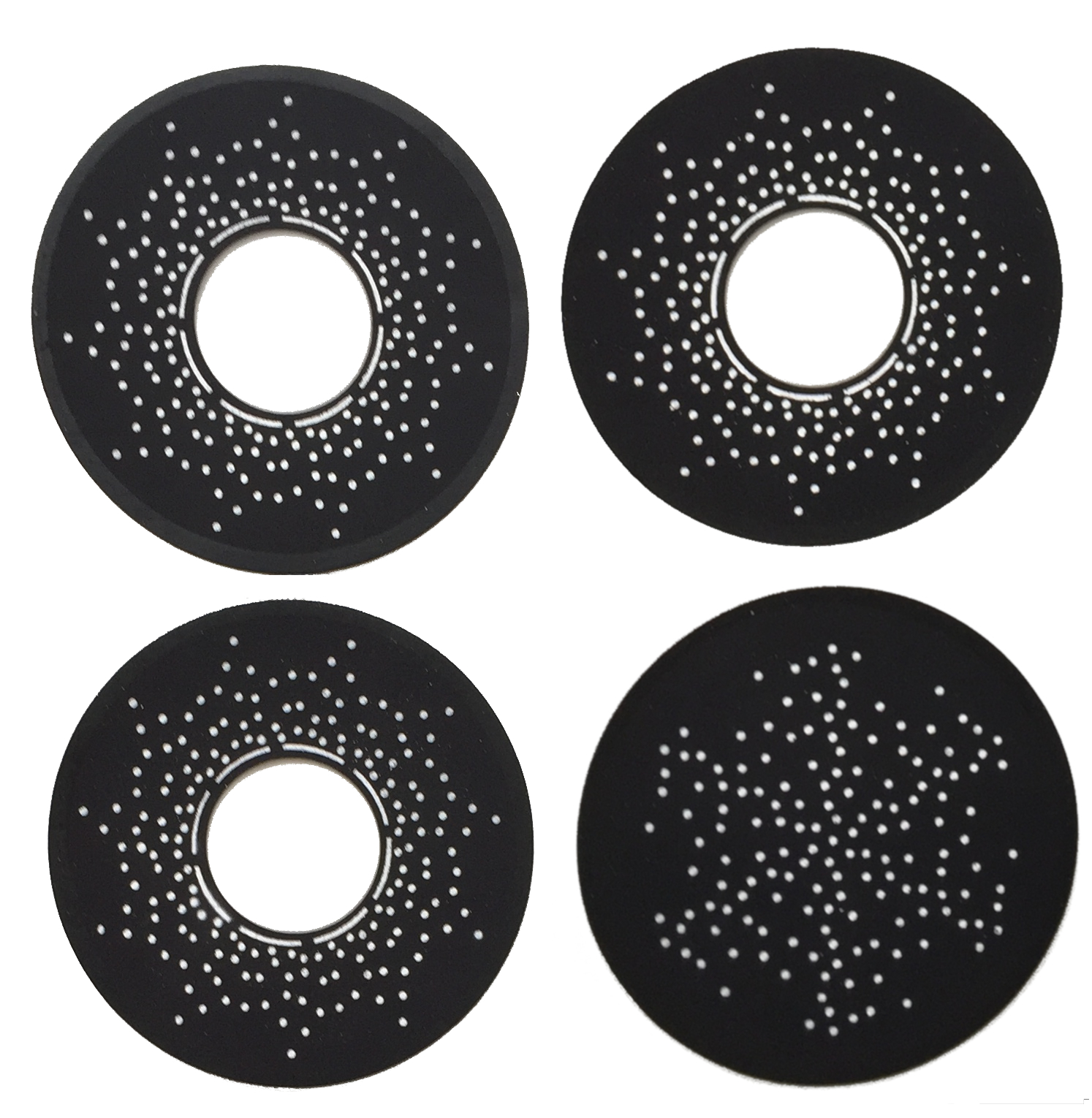}
	\caption{Pattern applied to parts of the actuator visible to the integrated camera before assembly. White dashed rings are included around the circular cut-outs to make the individual bellows easily distinguishable from the camera's perspective. Additional dots (\unit[2]{mm} diameter) are scattered around these rings to observe the displacement of the bellows. }
	\vspace{-0.5cm}
	\label{Fig:Pattern}
\end{figure}

Before assembly, a pattern is applied to the fabric layers on the interior of the actuator that are visible to the integrated camera. This is done to provide the camera with visual features to track. The pattern is applied with white textile spray paint (319921 textile spray paint by Dupli-Color) to provide a high contrast to the black fabric. First, the pattern is cut from adhesive stencil film (S380 by ASLAN) with a laser cutter. In a second step, the stencil is attached to the relevant fabric layers of the bellows and the pattern is applied in four successive, light coats. It is important to keep the paint layers thin, to prevent them from smearing in the heat press during assembly. The applied pattern can be seen in Fig. \ref{Fig:Pattern}. 

\begin{figure}
	\centering
	\begin{minipage}[b]{0.59\columnwidth}
		\includegraphics[width=\columnwidth]{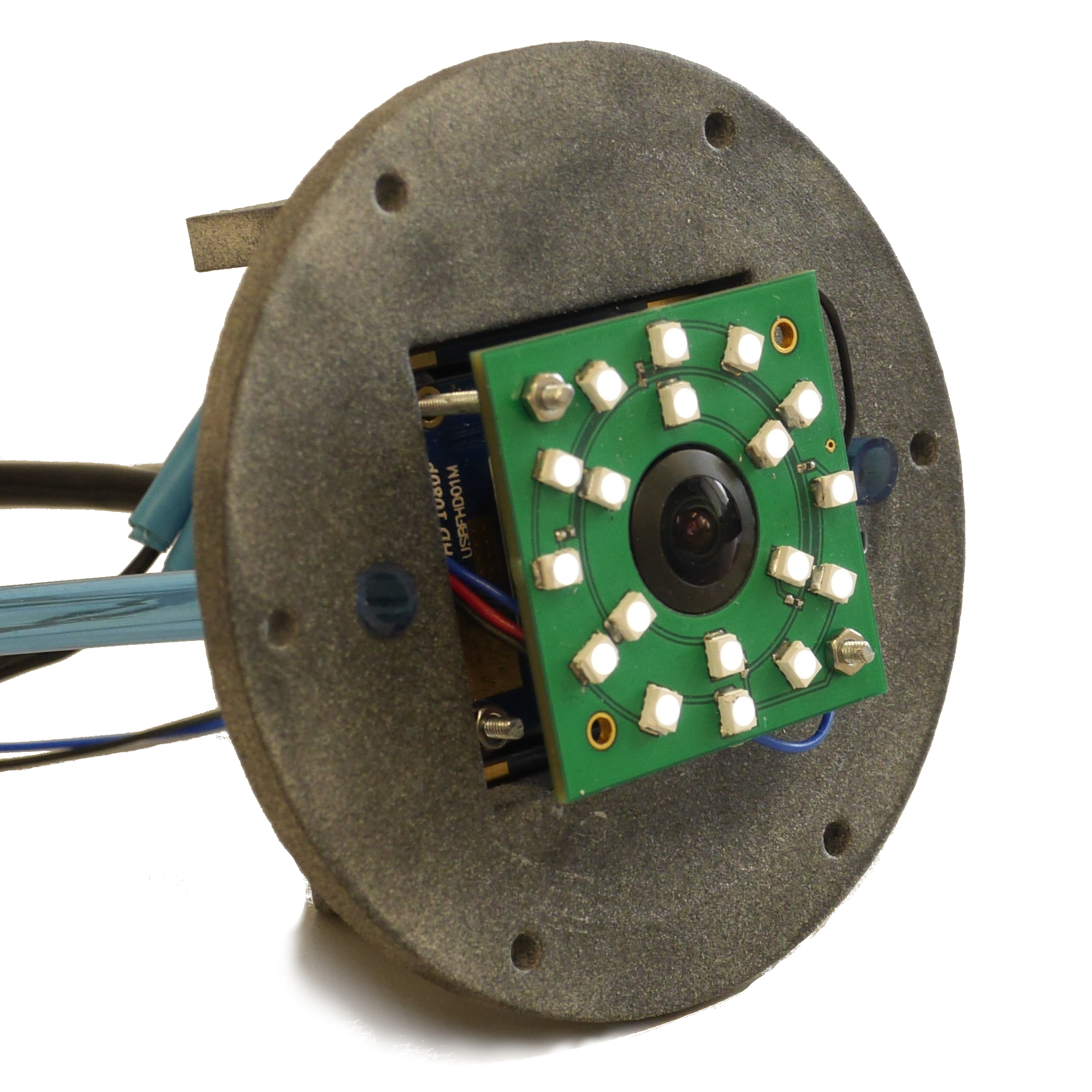}	
	\end{minipage}
	\begin{minipage}[b]{0.39\columnwidth}
		\includegraphics[width=0.9\columnwidth]{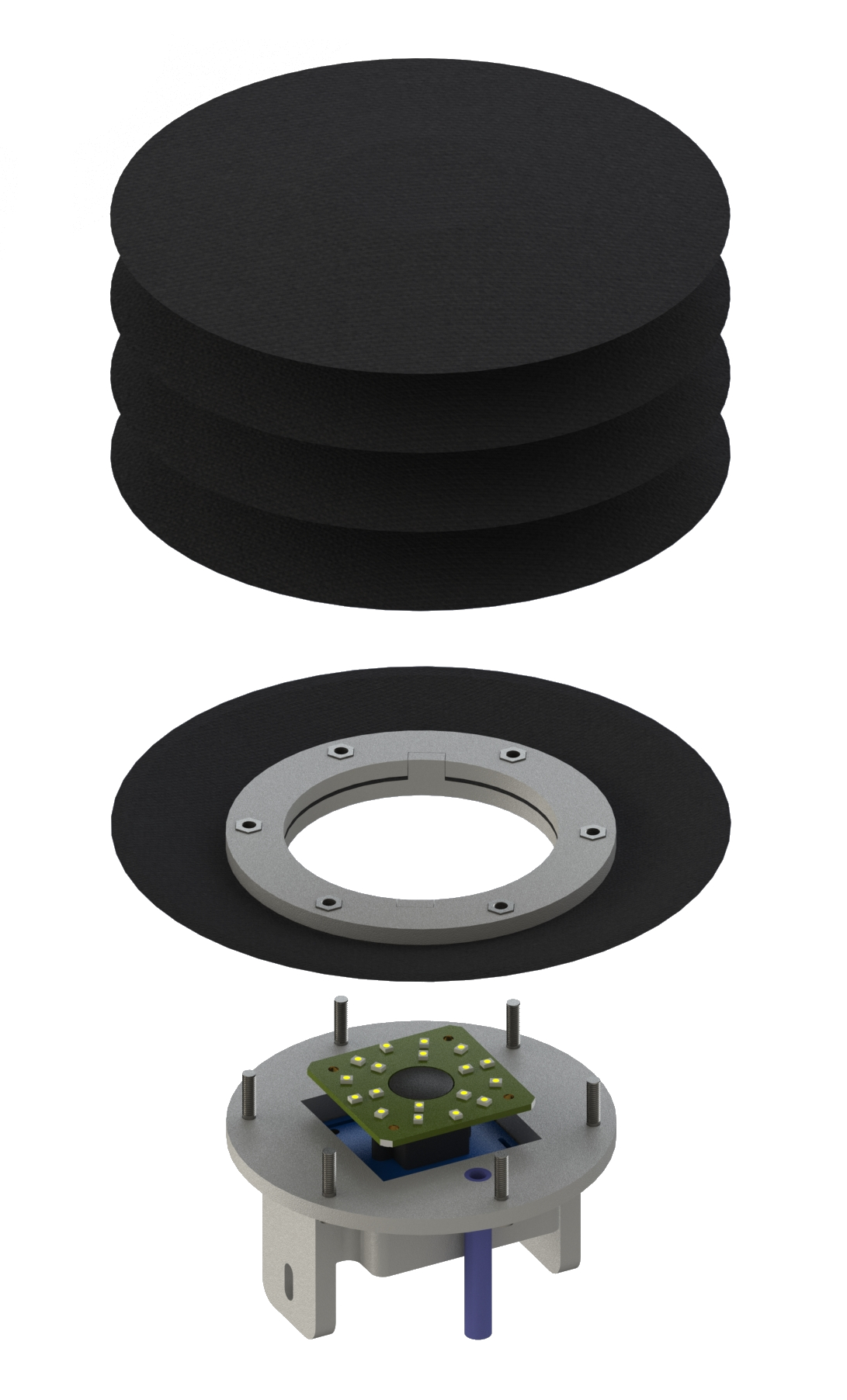}	
	\end{minipage}
	\caption{Left: Camera fixture that is screwed to the bottom of the actuator. It features a USB camera with a 180$^\circ$ fisheye lens and a LED board to illuminate the interior of the actuator. Air to drive the actuator is supplied through the blue tubing. Right: Rendering of the actuator in  exploded view. A two-part 3D printed flange is glued to the opening of the first bellow. The camera fixture is secured to it with six screws to ensure airtightness. When assembled, the camera does not protrude beyond the flange.}
	\vspace{-0.3cm}
	\label{Fig:Camera}
\end{figure}

A 3D printed flange (made from PA12, as all 3D printed parts) is glued to the first bellow (using Loctite 4850) as an interface for the camera. The camera is attached to a separate 3D printed fixture that is connected to the flange of the actuator with six screws to ensure airtightness, see Fig. \ref{Fig:Camera}. Pressurized air is supplied to the actuator through two blue hoses that are glued to openings in the camera fixture (using Loctite 4850).

The camera used (USBFHD01M-L180 by ELP) has a 180$^\circ$ fisheye lens, a resolution of 640x480 and can provide up to 100 frames per second. A LED board is placed around the lens to control the lighting and illuminate the pattern in the actuator. Positioning this board on the same plane as the lens ensures that potential shadows are eliminated. 

Aside from the camera, a time-of-flight distance sensor (VL6180X by STMicroelectronics) is integrated into a second actuator for comparison. This actuator is constructed identically to the first one, with the difference that the camera and LED board are replaced with the time-of-flight sensor. To improve the convergence time of the range measurements, a piece of reflective tape (Scotchlite 7610 by 3M) is placed on the interior of the last bellow facing the time-of-flight sensor.

\subsection{Test Setup}\label{ssec:pattern}
The test setup for the actuators includes all required components to control their internal pressure and the motion capture system used as ground truth. 

Air can be supplied with two different methods. Manually, using a pressure regulator and with an automated pressure control system. For the automated pressure control of the actuator, fast switching valves (MHJ10 by Festo) are used. These valves are operated with pulse width modulation to achieve continuous airflow. To increase the performance of the pressure control, a bypass is included. A more detailed description of the full pressure control approach including the bypass can be found in \cite{hofer2018design}. 
The pressure control system and the time-of-flight sensor are interfaced with a LabJack T7 Pro.

A motion capture system (using T40-S cameras by Vicon) with sub-millimeter accuracy is employed to obtain ground truth data at $\unit[200]{Hz}$. 
\section{Method}\label{sec:method}

\begin{figure*}
	\centering
	\includegraphics[width=1\textwidth]{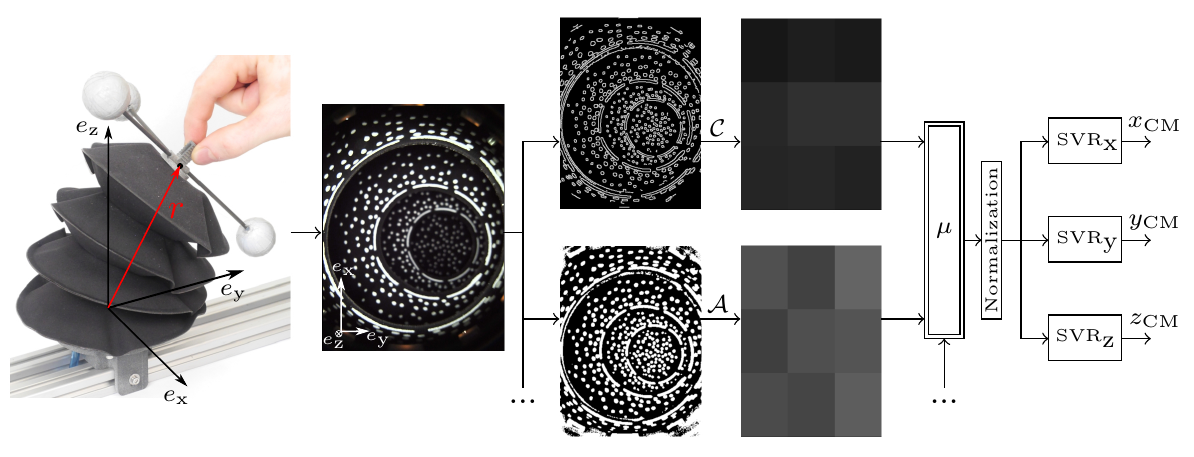}

	\caption{Illustration of how the machine learning pipeline predicts the position of the point of interest $r$ from the image captured by the integrated camera. First features are extracted from images by applying an array of image filters and averaging the pixel values of the resulting images over a rectangular grid. The resulting intensity values are then concatenated yielding a feature vector $\mu$. Two of the filters (Canny edge detection resulting in $\mathcal{C}$ and adaptive thresholding resulting in $\mathcal{A}$) are visualized above. The location $r$ is then predicted by evaluating three SVR models on $\mu$, to predict its components separately. The subscript CM denotes predictions from the camera-based pipeline. }
	\vspace{-0.3cm}
	\label{Fig:Pipeline}
\end{figure*}

The goal of the proposed sensing approach is to reconstruct the 3D position of a point of interest on the actuator using only images from the integrated camera. This point is located in the center of the marker frame on the actuator. An inertial Cartesian coordinate system $I$ is introduced with the origin $O$ directly above the camera lens. Let $r$ denote the vector pointing from $O$ to the point of interest and $_Ir = (x,y,z) \in \mathbb{R}^3$ being its components in the inertial coordinate frame $I$, see Fig. \ref{Fig:Pipeline}.

The task of reconstructing $r$ is solved in two steps. First, a set of features are computed from the image. In a second step, $r$ is predicted by evaluating SVR models on the extracted features. 

The remainder of this section is structured as follows. A method for extracting features from the images is presented in Subsection \ref{ssec:Features}. The SVR is discussed in Subsection \ref{ssec:SVR}. The data collection and training of the SVR models are discussed in Subsections \ref{ssec:Data} and \ref{ssec:Modellearning}.

\subsection{Feature Extraction}\label{ssec:Features}
\begin{figure}
	\centering
	\begin{tikzpicture}

\draw  (-1.25,1) rectangle (-0.1,0) node[pos=.5]{Image};
\draw  (-1.2,0.95) rectangle (-0.15,0.05) ;
\draw  (0.15,2.5) rectangle (0.65,-1.5) node[pos = .5,rotate=90]{Grayscale Conversion};
\draw[->] (-0.1,0.5) -- (0.15,0.5);
\draw  (1.2,0.25) rectangle (4.05,-0.25) node [pos=.5] {Ad. Thresholding};
\draw  (1.2,2) rectangle (4.05,1.5)node [pos=.5] {Bin. Thresholding};
\draw  (1.2,3) rectangle (4.05,2.5)node [pos=.5] {Canny Edge Det.};
\draw  (1.2,-1.75) rectangle (4.05,-2.25)node [pos=.5] {Dilation};
\draw  (1.2,-0.75) rectangle (4.05,-1.25)node [pos=.5] {Erosion};
\draw (0.65,0.5) -- (1,0.5) node [above,midway] {$\mathcal{G}$} -- (1,2.75);

\draw (1,0) -- (1,0.5);
\draw[->] (1,2.75) -- (1.2,2.75);
\draw[->] (1,1.75) -- (1.2,1.75);
\draw  (1.2,1.3) rectangle (3.2,0.9)node [pos=.5]{\small{Averaging}};
\draw[->] (2.2,1.3) -- (2.2,1.5);
\draw[->] (1,0.5) -- (4.65,0.5);
\draw[->] (2.2,0.5) -- (2.2,0.9);
\draw  (3.45,0.9) rectangle (4.05,1.3) node [pos=.5]{$\scriptstyle {b_\mathcal{M}}$};
\draw[->] (3.75,1.3) -- (3.75,1.5);
\draw[->] (4.05,2.75) -- (4.65,2.75) node [above,midway]{$\mathcal{C}$};
\draw[->] (4.05,1.75) -- (4.65,1.75) node [above,midway]{$\mathcal{M}$};
\draw[->] (4.05,0) -- (4.65,0) node [above,midway]{$\mathcal{A}$};
\draw[->] (4.05,-1) -- (4.65,-1) node [above,midway]{$\mathcal{E}$};
\draw[->] (4.05,-2) -- (4.65,-2) node [above,midway]{$\mathcal{D}$};

\draw[->] (6.05,1.5) -- (6.8,1.5) node [above,midway]{$\mu_\mathcal{C}$};
\draw[->] (6.05,1) -- (6.8,1) node [above,midway]{$\mu_\mathcal{M}$};
\draw[->] (6.05,0) -- (6.8,0) node [above,midway]{$\mu_\mathcal{A}$};
\draw[->] (6.05,-0.5) -- (6.8,-0.5) node [above,midway]{$\mu_\mathcal{E}$};
\draw[->] (6.05,-1) -- (6.8,-1) node [above,midway]{$\mu_\mathcal{D}$};
\draw[->] (6.05,0.5) -- (6.8,0.5) node [above,midway]{$\mu_\mathcal{G}$};

\draw (4.35,0) -- (4.35,-0.5) -- (1,-0.5) -- (1,-2);
\draw[->] (1,-2) -- (1.2,-2);
\draw[->] (1,-1) -- (1.2,-1);
\draw[->] (1,0) -- (1.2,0);
\draw  (4.65,3.25) rectangle (5.15,-2.5) node[pos = .5,rotate=90]{Average Pooling};
\draw  (5.55,3.1) rectangle (6.05,-2.35) node[pos = .5,rotate=90]{Flattening};

\draw[->] (5.15,0) -- (5.55,0);
\draw[->] (5.15,0.5) -- (5.55,0.5);
\draw[->] (5.15,-2) -- (5.4,-2) -- (5.4,-1) -- (5.55,-1);
\draw[->] (5.15,-1) -- (5.3,-1) -- (5.3,-0.5) -- (5.55,-0.5);
\draw[->] (5.15,2.75) -- (5.4,2.75) -- (5.4,1.5) -- (5.55,1.5);
\draw[->] (5.15,1.75) -- (5.3,1.75) -- (5.3,1) -- (5.55,1);
\draw[->] (4.9,3.5) -- (4.9,3.25);
\draw  (4.65,3.5) rectangle (5.15,4) node [pos=.5] {$S$};
\draw  (6.8,2) rectangle (7.35,-1.55)  node [pos=.5]  {$ \mu$};
\draw  (6.85,1.95) rectangle (7.3,-1.5);
\end{tikzpicture}
	\vspace{-0.5cm}
	\caption{Before predicting $r$ with kernelized SVR, the image data is compressed. This is done by converting the images to grayscale and applying a series of image filters. The resulting images are averaged across an $S\times S$ grid of evenly-sized rectangular regions (average pooling) and the resulting intensity values concatenated to the feature vector $\mu$. }
	\vspace{-0.3cm}
	\label{Fig:FeatureExtraction}
\end{figure}
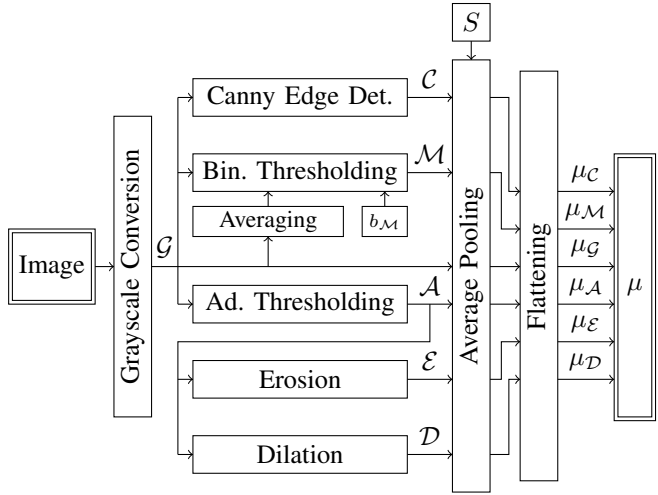

The following procedure is employed to obtain the features from a single image:
First, the image is converted to a grayscale image $\mathcal{G}$. In a second step, $\mathcal{G}$ is passed through an array of five image filters producing a total of six images including $\mathcal{G}$. Fig. \ref{Fig:FeatureExtraction} illustrates how these filters are applied. An adaptive thresholding filter is applied to $\mathcal{G}$ producing $\mathcal{A}$. The result is then dilated and eroded, yielding $\mathcal{D}$ and $\mathcal{E}$. A Canny edge detector is applied to $\mathcal{G}$ resulting in $\mathcal{C}$. The last image $\mathcal{M}$ is obtained by using a binary thresholding filter on $\mathcal{G}$. The threshold of this filter is chosen to be the average of the pixel values of $\mathcal{G}$, denoted by $\bar{\mu}_\mathcal{G}$, plus an offset $b_{ \mathcal{M}}$. The features are then computed by splitting every image into $S \times S$ evenly-sized rectangular regions and averaging the pixel values across these regions (average pooling), where $S$ is treated as a hyperparameter. The feature vector $\mu \in \mathbb{R}^{6S\cdot S}$ is then obtained by concatenating these averaged intensity values.
The OpenCV\footnote{\url{https://opencv.org/}} implementation of the above-mentioned filters is used with the parameters listed in the Appendix.

\subsection{Support Vector Regression}\label{ssec:SVR}
To find a mapping between the features $\mu$ and $r$, kernelized support vector regression with a radial basis function (RBF) kernel is used. This is done by using three regressors to predict the three components of $r$ separately. Every regressor has three hyperparameters that need to be tuned. Namely: $\epsilon$, the parameter that controls the $\epsilon$-insensitive loss function, $K$, the weighting factor that determines the relative cost between the loss function and the $\textit{L}_2$-Regularization, and $\gamma$ the kernel parameter that determines the width of the RBF. More details on support vector regression can be found in \cite{smola2004tutorial}. The employed hyperparameters are summarized in Section \ref{ssec:Modellearning}.

\subsection{Data Collection}\label{ssec:Data}

To evaluate, train and tune the proposed pipeline, multiple image data sets are collected. This is done by manually inflating the actuator and moving the grip in $e_\text{x}$ and $e_\text{y}$ directions (see Fig. \ref{Fig:Pipeline}) while simultaneously recording the images from the integrated camera. The images are saved at 10 Hz together with the corresponding ground truth data from the motion capture system. Using this method, around 9000 training images and 500 additional test images are recorded. The process of collecting the necessary image data takes around 20 minutes.

\subsection{Model Learning}\label{ssec:Modellearning}

To obtain the hyperparameters of the $x$, $y$ and $z$-regressors for a given $S$, 5-fold cross-validation is performed on the training images using a grid-search. Since $S$ directly steers the dimension of $\mu$, and therefore the number of parameters of the models, only values from one to four are considered.

Before training, the individual entries of $\mu$ are normalized by subtracting the mean and dividing by the empirical standard deviation. The training of the SVR models is performed in Python with the scikit-learn library\footnote{\url{https://scikit-learn.org/}}. Training a set of three regressors takes about 30 seconds (for $S=4$) on the employed laptop\footnote{Intel Core i7-8550U @ $\unit[1.80]{GHz}$}. This allows for an extensive search of the hyperparameter space.



It was observed on the image data, that generally the accuracy of the predictions improved when $S$ was increased.
The hyperparameters selected through cross-validation for $S=3$ are listed in Table \ref{Tab1:Hyp}. In Section \ref{sec:experiments}, $S = 3$ is used since it yields a good trade-off between computational cost and the accuracy of the models.

\begin{table}[h!]
	\caption{Hyperparameters for S=3}
	\label{Tab1:Hyp}
	
	\centering
	\begin{tabular}{c|ccc}
		Regressor&$\epsilon$&$K$&$\gamma$\\
		\hline
		$x$\Tstrut& 0.96 & 112.5 & 0.060\\
		$y$& 0.96 & 112.5 & 0.060 \\
		$z$& 1.00 & 189.0 & 0.005 \\
	\end{tabular}
    \vspace{-0.3cm}
\end{table}


\section{Experiments}\label{sec:experiments}
In this section the camera-based sensing approach is evaluated. A performance baseline is established with a time-of-flight distance sensor. The vision-based approach is then used in two different settings. First, the predictions of $r$ are computed in real-time while the actuator is moved in the same fashion as during training. In a second experiment, the proposed sensing approach is used for closed loop elongation control of the actuator. The root-mean-squared error ($Rmse$) is used as the evaluation metric for the experiments. In the following the subscript GT refers to ground truth data from the motion capture system.

\subsection{Performance Baseline}\label{ssec:ToF}
\begin{figure}
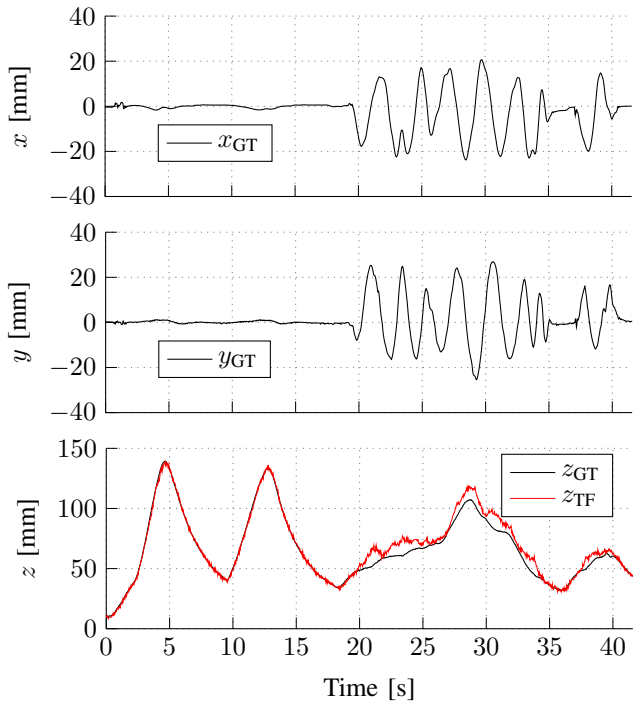

	\centering 
	\input{img/TOFx}
	\input{img/TOFy}
	\input{img/TOFz}
	\vspace{-0.5cm}
	\caption{ A time-of-flight distance sensor is integrated into an actuator and calibrated to measure the $z$-component of $r$. The calibrated measurements, $z_\text{TF}$, are compared to data from the motion capture system ($z_\text{GT}$). Two scenarios can be observed in the plots. First the actuator is left undisturbed. After 19 seconds the grip on the actuator is manually moved in $e_\text{x}$ and $e_\text{y}$ directions. The $Rmse$ in $z$-direction is  $\unit[1.37]{mm}$ on the first part of the trajectory (undisturbed) and $\unit[6.19]{mm}$ on the remainder of the trajectory (disturbed).}
	\vspace{-0.5cm}
	\label{Fig:TOF}
\end{figure}

The time-of-flight sensor is chosen as reference, because it presents a straightforward solution to measuring the elongation of a linear actuator in a non-interacting fashion. The sensor is first calibrated by finding an approximate mapping from the raw sensor readings to the ground truth motion capture data using linear regression. The calibrated measurements from the time-of-flight (TF) sensor are denoted by $z_\text{TF}$. The actuator is not additionally moved in $e_\text{x}$ and $e_\text{y}$ directions during calibration since the sensor can only measure distance, without discerning between horizontal and vertical displacement. The resulting calibration is then evaluated on a trajectory shown in Fig. \ref{Fig:TOF} with and without lateral motion. The sensor produces an $Rmse$ of $\unit[1.37]{mm}$ when the actuator is not laterally moved (up to 19 seconds on the trajectory mentioned). As expected, it can clearly be seen that if the actuator is moved laterally (after 19 seconds on the trajectory), the performance degrades significantly yielding an $Rmse$ of $\unit[6.19]{mm}$.

\subsection{Real-Time Prediction}\label{ssec:RTpred}
\begin{figure}
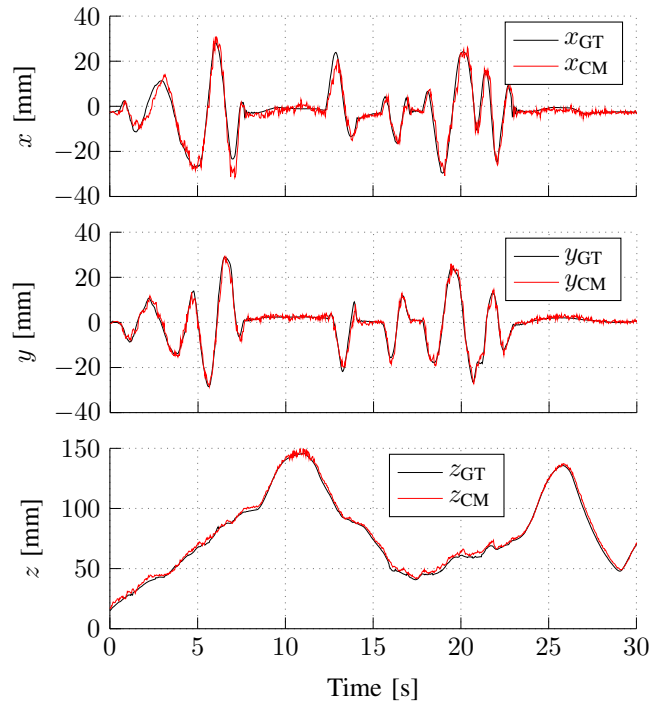

	\centering 
	\input{img/V3RTpredx}
	\input{img/V3RTpredy}
	\input{img/V3RTpredz}
	\vspace{-0.5cm}
	\caption{Prediction of $r$ computed at 40 Hz from camera images while the actuator is moved laterally. The predictions are plotted against the measurements from the motion capture system (subscript GT). The $Rmse$ in the individual components are $\unit[2.77]{mm}$ in $x$, $\unit[1.65]{mm}$ in $y$ and $\unit[2.33]{mm}$ in $z$-direction.}
	\vspace{-0.5cm}
	\label{Fig:RTpred}
\end{figure}

To observe the performance and the real-time capability of the camera-based sensing approach, the actuator is manually inflated and moved in the same fashion as during training (see Fig. \ref{Fig:SoftActuator}). The resulting performance on a sample trajectory is shown in Fig. \ref{Fig:RTpred}. Predictions from the camera-based pipeline are denoted by the subscript CM. The hyperparameters stated in Subsection \ref{ssec:Modellearning} are used for training since selecting $S=3$ is a good trade-off between accuracy and computational cost of evaluating the pipeline.

The resulting sensing pipeline runs reliably at $\unit[40]{Hz}$ using the same laptop as for training. The $Rmse$ in the individual components are $\unit[2.77]{mm}$ in $x$, $\unit[1.65]{mm}$ in $y$ and $\unit[2.33]{mm}$ in $z$-direction. It can be seen that the camera-based approach circumvents the performance degradation, when the actuator is moved laterally.
Note that the error in the $x$-direction is significantly larger than in the $y$-direction. This issue presumably arises because of asymmetries in the pattern or because of the rectangular shape of the image and requires further investigation. 

The performance is considerably reduced if the actuator is elongated less than $\unit[20]{mm}$. Upon inspection of training images recorded at or below this elongation, it is seen that the lighting conditions are drastically different to  the images recorded at elongations above $\unit[20]{mm}$. This issue is a current limitation of the approach. 

\subsection{Elongation Control}
The elongation of the actuator is controlled using the camera-based predictions as sensory feedback. This is done by using a cascaded control structure that separates the faster pressure dynamics from the elongation dynamics of the actuator. A proportional-integral (PI) position controller is used to output a pressure setpoint to an automated pressure control system. The implementation of the pressure controller is outlined in \cite{hofer2018design}. Air lost through the bypass is accounted for by augmenting the pressure controller with a second-order polynomial feedforward element that is tuned manually.
 
\begin{figure}
	\centering 
	\input{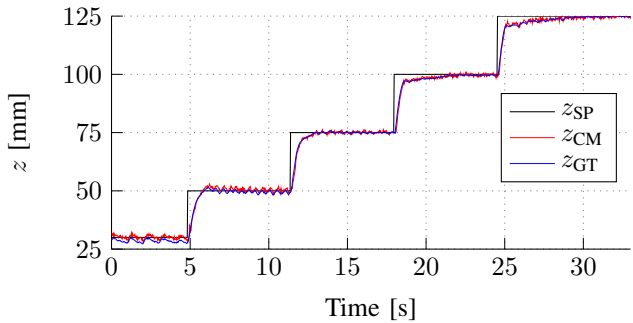}
	\vspace{-0.2cm}
	\caption{Elongation control using camera-based sensing for feedback. The $z$-component $z_\text{CM}$ of the position $r$ is predicted from camera images at $\unit[50]{Hz}$ and used for feedback control. Note that $z_\text{SP}$ denotes the setpoint trajectory.  The $Rmse$ between $z_\text{GT}$ and the prediction $z_\text{CM}$ is \unit[1.01]{mm}.}
	\vspace{-0.4cm}
	\label{Fig:ElCtrl}
\end{figure}

The entire pipeline and the controllers are implemented in C++. The position controller is executed in the same thread as the camera-based sensing pipeline, which runs at $\unit[50]{Hz}$ (the predictions of the $x$ and $y$-components of $r$ are disabled to increase the sampling rate). The pressure controller is executed in a separate thread running at $\unit[100]{Hz}$. Fig. \ref{Fig:ElCtrl} shows the system tracking a series of elongation steps. 
It can be seen that the setpoint trajectory can be reliably tracked with the camera-based sensing pipeline for feedback.
The $Rmse$ between $z_\text{CM}$ and $z_\text{GT}$ is $\unit[1.01]{mm}$, which is close to the performance of the time-of-flight sensor, when the actuator is not laterally perturbed. 

Nevertheless, it is observed that using the proposed camera-based sensing approach in combination with the switching valves induces oscillations in $z$ for lower elongation ranges. This can be seen in Fig. \ref{Fig:ElCtrl}, for example when $z_\text{SP}$ is $\unit[50]{mm}$. To reach this elongation, an internal pressure of approximately $\unit[0.002]{bar}$ above ambient pressure is required. Controlling such small pressure changes is challenging with the employed switching valves and required careful tuning of the controller including the bypass approach. Using $z_{GT}$ for feedback instead, reduces the amplitude of the oscillations from approximately $\unit[2]{mm}$ to $\unit[1]{mm}$. The same holds for the first setpoint. This limitation requires further investigation.

\section{Conclusion}\label{sec:conclusion}
A camera-based sensing approach for an air-driven linear soft actuator has been presented as a proof of concept. The proposed sensing pipeline first extracts features from images generated with an integrated camera using classical image filters and average pooling. The resulting average intensity values are then used as input features for three SVR models that jointly predict the position of the point of interest $r$, on the actuator. 
The proposed approach performs similarly to a performance baseline made with an off-the-shelf distance sensor. Additionally, the camera-based approach benefits from the rich visual information of the pattern, which allows it to predict the 3D position of the point of interest. Moreover, it was demonstrated that the pipeline presented can successfully be used for closed loop elongation control of the actuator.

Future work will include investigating whether this approach can be scaled to track multiple points simultaneously, its generalization to other actuator geometries and exploring different pattern-feature combinations.

\section*{Acknowledgment}
The authors would like to thank Michael Egli for his contribution to the development of the prototype.
\section*{Appendix}
\subsection{Filter Parameters}\label{appendix:Filtparam}
The OpenCV frame work was used for the image processing. Table \ref{Table:FilterParameters} contains the parameters of the filters employed.
Note that $\bar{\mu}_\mathcal{G}$ is the average across all entries of $\mu_\mathcal{G}$. This value plus the offset $b_\mathcal{M}$ is chosen as the threshold for a binary thresholding filter.

\begin{table}[h!]
	\caption{OpenCV Filter Parameters}
	\centering
	\small
	
	\begin{tabular}{ p{3cm}p{2cm}p{1.5cm}}
		\hline
		\hline
		Filter&Parameter&Value\Tstrut\\
		\hline
		\hline
		Adaptive Thresholding\Tstrut&\multirow{5}{2cm}{maxValue\\adaptiveMethod\\thresholdType\\Blocksize\\C}&\multirow{5}{1.5cm}{255\\Gaussian\\Binary\\57\\2}\\
		&&\\
		&&\\
		&&\\
		&&\\
		\hline
		
		Binary Thresholding\Tstrut&\multirow{3}{2cm}{maxValue\\$b_\mathcal{M}$\\Threshold\\Type}&\multirow{3}{1.5cm}{255\\100\\$\bar{\mu}_\mathcal{G}+b_\mathcal{M}$\\Binary}\\
		&&\\
		&&\\
		&&\\
		\hline
		
		Canny Edge Detection\Tstrut&\multirow{3}{2cm}{lowThreshold\\highThreshold\\kernelSize}&\multirow{3}{1.5cm}{100\\130\\3}\\
		&&\\
		&&\\
		\hline
		
		Dilation\Tstrut&\multirow{2}{2cm}{kernelShape\\kernelSize}&\multirow{2}{1.5cm}{square\\5x5} \\
		&&\\
		\hline
		
		Erosion\Tstrut&\multirow{2}{2cm}{kernelShape\\kernelSize}&\multirow{2}{1.5cm}{square\\5x5} \\
		&&\\
		\hline
	\end{tabular}
	\label{Table:FilterParameters}
	\vspace{-0.5cm}
\end{table}
\bibliographystyle{IEEEtran}
\bibliography{bibliography}

\begin{thebibliography}{10}
\providecommand{\url}[1]{#1}
\csname url@samestyle\endcsname
\providecommand{\newblock}{\relax}
\providecommand{\bibinfo}[2]{#2}
\providecommand{\BIBentrySTDinterwordspacing}{\spaceskip=0pt\relax}
\providecommand{\BIBentryALTinterwordstretchfactor}{4}
\providecommand{\BIBentryALTinterwordspacing}{\spaceskip=\fontdimen2\font plus
\BIBentryALTinterwordstretchfactor\fontdimen3\font minus
  \fontdimen4\font\relax}
\providecommand{\BIBforeignlanguage}[2]{{%
\expandafter\ifx\csname l@#1\endcsname\relax
\typeout{** WARNING: IEEEtran.bst: No hyphenation pattern has been}%
\typeout{** loaded for the language `#1'. Using the pattern for}%
\typeout{** the default language instead.}%
\else
\language=\csname l@#1\endcsname
\fi
#2}}
\providecommand{\BIBdecl}{\relax}
\BIBdecl

\bibitem{PPolygerinos_SRR}
P.~Polygerinos, N.~Correll, S.~A. Morin, B.~Mosadegh, C.~D. Onal, K.~Petersen,
  M.~Cianchetti, M.~T. Tolley, and R.~F. Shepherd, ``{Soft Robotics: Review of
  Fluid-Driven Intrinsically Soft Devices; Manufacturing, Sensing, Control, and
  Applications in Human-Robot Interaction},'' \emph{Advanced Engineering
  Materials}, vol.~19, no.~12, p. 1700016, 2017.

\bibitem{DRus_Review}
D.~Rus and M.~Tolley, ``Design, fabrication and control of soft robots,''
  \emph{Nature}, vol. 521, pp. 467--475, 2015.

\bibitem{HWang_TowardsPerceptiveSR}
H.~Wang, M.~Totaro, and L.~Beccai, ``{Toward Perceptive Soft Robots: Progress
  and Challenges},'' \emph{Advanced Science}, vol.~5, no.~9, p. 1800541, 2018.

\bibitem{HZhao_Waveguide}
H.~Zhao, K.~O{\textquoteright}Brien, S.~Li, and R.~F. Shepherd,
  ``Optoelectronically innervated soft prosthetic hand via stretchable optical
  waveguides,'' \emph{Science Robotics}, vol.~1, no.~1, 2016.

\bibitem{MDobrzynski_Contacless}
M.~K. {Dobrzynski}, R.~{Pericet-Camara}, and D.~{Floreano}, ``Contactless
  deflection sensor for soft robots,'' in \emph{2011 IEEE/RSJ International
  Conference on Intelligent Robots and Systems (IROS)}, 2011.

\bibitem{HYang_ModelAndAnalysis}
H.~D. Yang, B.~T. Greczek, and A.~T. Asbeck, ``{Modeling and Analysis of a
  High-Displacement Pneumatic Artificial Muscle With Integrated Sensing},''
  \emph{Frontiers in Robotics and AI}, vol.~5, p. 136, 2019.

\bibitem{CChorley_TacTip}
C.~Chorley, C.~Melhuish, T.~Pipe, and J.~Rossiter, ``{Development of a Tactile
  Sensor Based on Biologically Inspired Edge Encoding},'' \emph{Advanced
  Robotics, ICAR}, 2009.

\bibitem{CSferrazza_DesignMot}
C.~Sferrazza and R.~D'Andrea, ``{Design, Motivation and Evaluation of a
  Full-Resolution Optical Tactile Sensor},'' \emph{Sensors}, vol.~19, no.~4,
  2019.

\bibitem{WYuan_GelSight}
W.~Yuan, S.~Dong, and E.~H. Adelson, ``{GelSight: High-Resolution Robot Tactile
  Sensors for Estimating Geometry and Force},'' \emph{Sensors}, vol.~17,
  no.~12, 2017.

\bibitem{ZKappassov_TactileSensing}
Z.~Kappassov, J.-A. Corrales, and V.~Perdereau, ``Tactile sensing in dexterous
  robot hands,'' \emph{Robotics and Autonomous Systems}, vol.~74, pp. 195--220,
  2015.

\bibitem{BMcInroe_TowardsAS}
B.~W. McInroe, C.~L. Chen, K.~Y. Goldberg, R.~Bajcsy, and R.~S. Fearing,
  ``{Towards a Soft Fingertip with Integrated Sensing and Actuation},'' in
  \emph{2018 IEEE/RSJ International Conference on Intelligent Robots and
  Systems (IROS)}, 2018.

\bibitem{RWang_RealTimeSoftRobot}
R.~Wang, S.~Wang, E.~Xiao, K.~Jindal, W.~Yuan, and C.~Feng, ``{Real-time Soft
  Robot 3D Proprioception via Deep Vision-based Sensing},'' vol.
  abs/1904.03820, 2019.

\bibitem{yang2018new}
H.~D. Yang and A.~T. Asbeck, ``{A New Manufacturing Process for Soft Robots and
  Soft/Rigid Hybrid Robots},'' in \emph{2018 IEEE/RSJ International Conference
  on Intelligent Robots and Systems (IROS)}, 2018.

\bibitem{hofer2018design}
M.~Hofer and R.~D'Andrea, ``{Design, Modeling and Control of a Soft Robotic
  Arm},'' in \emph{2018 IEEE/RSJ International Conference on Intelligent Robots
  and Systems (IROS)}, 2018.

\bibitem{smola2004tutorial}
A.~J. Smola and B.~Sch{\"o}lkopf, ``A tutorial on support vector regression,''
  \emph{Statistics and computing}, vol.~14, no.~3, 2004.

\end{thebibliography}

\end{document}